\documentclass[10pt,twocolumn,letterpaper]{article}

\usepackage[pagenumbers]{cvpr}      

\usepackage{graphicx}
\usepackage{amsmath}
\usepackage{amssymb}
\usepackage{makecell}
\usepackage{times}
\usepackage{float}
\usepackage{dsfont}
\usepackage{comment}
\usepackage{amsmath,amssymb} 
\usepackage{color}
\usepackage{caption}
\usepackage{cite}
\usepackage{epsfig}
\usepackage{mathtools}
\usepackage{tabularx}
\usepackage{booktabs}
\usepackage{arydshln}
\usepackage{multirow}
\usepackage{enumitem}
\usepackage{wrapfig}
\usepackage{bm}
\usepackage{algorithm}
\usepackage{algpseudocode}
\usepackage[nopar]{lipsum}
\usepackage{arydshln}
\usepackage{pifont} 

\usepackage[table]{xcolor}
\usepackage{booktabs}
\usepackage{colortbl}
\definecolor{Gray}{gray}{0.9}

\usepackage[pagebackref,breaklinks,colorlinks]{hyperref} 
\usepackage[accsupp]{axessibility}  

\usepackage[accsupp]{axessibility}
\usepackage[capitalize]{cleveref}
\crefname{section}{Sec.}{Secs.}
\Crefname{section}{Section}{Sections}
\Crefname{table}{Table}{Tables}
\crefname{table}{Tab.}{Tabs.}

\def\cvprPaperID{5821} 
\def\confName{CVPR}
\def\confYear{2023}

\DeclareMathAlphabet\mathbfcal{OMS}{cmsy}{b}{n}

\def\orange#1{\textcolor[rgb]{1,0.49,0}{#1}}

\definecolor{deepGreen}{RGB}{0,153,0}
\definecolor{orange}{RGB}{255,125,0}

\newcommand{\scriptveryshortarrow}[1][3pt]{{%
    \hbox{\rule[\scriptratio\dimexpr\fontdimen22\textfont2-.2pt\relax]
               {\scriptratio\dimexpr#1\relax}{\scriptratio\dimexpr.4pt\relax}}%
   \mkern-4mu\hbox{\let\f@size\sf@size\usefont{U}{lasy}{m}{n}\symbol{41}}}}

\newcommand{\cut}[1]{}

\newcommand{\keypoint}[1]{\vspace{0.05cm}\noindent\textbf{#1}\;}  

\algnewcommand\algorithmicforeach{\textbf{for each}}
\algdef{S}[FOR]{ForEach}[1]{\algorithmicforeach\ #1\ \algorithmicdo}

\begin{document}
\title{Exploiting Unlabelled Photos for Stronger Fine-Grained SBIR}

\author{
Aneeshan Sain\textsuperscript{1,2}  \hspace{.2cm}
Ayan Kumar Bhunia\textsuperscript{1} \hspace{.3cm}
Subhadeep Koley\textsuperscript{1,2}  \hspace{.2cm}
Pinaki Nath Chowdhury\textsuperscript{1,2}  \hspace{.2cm}\\
Soumitri Chattopadhyay\thanks{Interned with SketchX}  \hspace{.2cm}
Tao Xiang\textsuperscript{1,2}\hspace{.2cm}  
Yi-Zhe Song\textsuperscript{1,2} \\
\textsuperscript{1}SketchX, CVSSP, University of Surrey, United Kingdom.  \\
\textsuperscript{2}iFlyTek-Surrey Joint Research Centre on Artificial Intelligence.\\
{\tt\small \{a.sain, a.bhunia, p.chowdhury,  t.xiang, y.song\}@surrey.ac.uk
} 
\vspace{-0.2cm}
}

\maketitle

\begin{abstract}
This paper advances the fine-grained sketch-based image retrieval (FG-SBIR) literature by putting forward a strong baseline that overshoots prior state-of-the-arts by $\approx$11\%. This is not via complicated design though, but by addressing two critical issues facing the community (i) the gold standard triplet loss does not enforce holistic latent space geometry, and (ii) there are never enough sketches to train a high accuracy model. For the former, we propose a simple modification to the standard triplet loss, that explicitly enforces separation amongst photos/sketch instances. For the latter, we put forward a novel knowledge distillation module can leverage photo data for model training. Both modules are then plugged into a novel plug-n-playable training paradigm that allows for more stable training. More specifically, {for (i) we employ an intra-modal triplet loss amongst sketches to bring sketches of the same instance closer from others, and one more amongst photos to push away different photo instances while bringing closer a structurally augmented version of the same photo (offering a gain of $\approx$4-6\%). To tackle (ii), we first pre-train a teacher on the large set of unlabelled photos over the aforementioned intra-modal photo triplet loss. Then we distill the contextual similarity present amongst the instances in the teacher's embedding space to that in the student's embedding space, by matching the distribution over inter-feature distances of respective samples in both embedding spaces (delivering a further gain of $\approx$4-5\%). Apart from outperforming prior arts significantly, our model also yields satisfactory results on generalising to new classes. Project page: \url{https://aneeshan95.github.io/Sketch_PVT/}}


\end{abstract}

\vspace{-0.7cm}
\section{Introduction}
\vspace{-0.2cm}
Sketch  \cite{chowdhury2023what,koley2023picture, bhunia2023sketch2saliency} has long established itself as a worthy query modality that is complementary to text \cite{dey2019doodle, dutta2019semantically, chowdhury2023scenetrilogy}. Conceived as a category-level retrieval task \cite{bui2018deep, collomosse2019query}, sketch-based image retrieval (SBIR) has recently taken a turn to a ``fine-grained’’ setting (i.e., FG-SBIR), where the emphasis is on fully utilising the faithful nature of sketches to conduct instance-level retrieval \cite{yu2016sketch, song2017deep}.

\begin{figure}
    \centering
    \includegraphics[width=\linewidth]{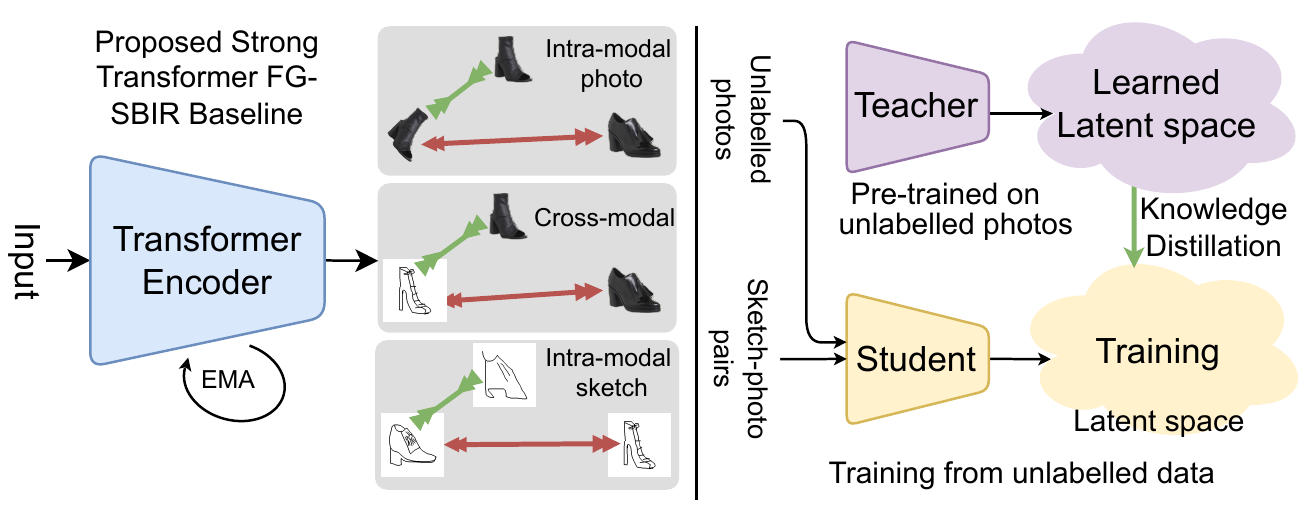} 
    \caption{{(left) A strong FG-SBIR baseline with strong PVT \cite{wang2021pyramid} backbone trained on intra-modal triplet loss, stabilised by EMA. (right) It additionally leverages unlabelled photos to enrich its latent space by distilling instance-wise discriminative knowledge of a teacher pre-trained on unlabelled photos.
    }}
    \label{fig:opener}
    \vspace{-0.7cm}
\end{figure}

Despite great strides made in the field, without exception, all existing FG-SBIR models \cite{yu2021fine, sain2023clip} work around a cross-modal triplet objective~\cite{yu2016sketch} to learn a discriminative embedding to conduct retrieval. The general intuition has always been to make a sketch sit closer to its paired photo while pushing away non-matching ones (\cref{fig:opener}). However, a conventional triplet setup does not enforce sufficient separation amongst different photos or sketch instances -- largely because the conventional objective {fails to retain the holistic latent space geometry, being overly restrictive on learning within-triplet feature separation}. 

For that and as our first contribution, we utilise an \textit{intra-modal} triplet objective in \textit{both} modalities, in addition to the regular \textit{cross-modal} \cite{yu2016sketch} triplet objective. For sketch, a query-sketch (anchor), is brought closer to another sketch of the same target-photo (positive) while distancing it from a sketch of any non-matching photo (negative). For the photo modality, as there is only one uniquely paired photo (anchor), we treat a morphologically augmented version of the photo as positive, and a non-matching one as negative. Importantly, we show that the best morphological operations are those that operate on visual attributes bearing relevance with information exclusive to sketches (e.g., shape) \cref{fig:opener}. 

The other, perhaps more pressing problem, facing the community is that of sketch data scarcity -- there are just not enough sketches for the extra performance gain \cite{bhunia2022sketching,bhunia2021vectorization}. Instead of going for the more explicit route of photo to pseudo sketch synthesis \cite{bhunia2021more}, we vote for the simple and more implicit route of leveraging unlabelled photos. For that, we adapt a knowledge distillation setup -- we first train a model on unlabelled \textit{photos only} via an intra-modal triplet loss, then distill its instance-wise discriminative knowledge to a FG-SBIR model. This apparently simple aim is however non-trivial to train owing to our cross-modal setup as naively using standard knowledge distillation paradigms involving logit distillation \cite{hinton2015distilling} from teacher's embedding space to the student's would be infeasible. For a cross-modal problem like ours, we need to preserve the instance-wise separation amongst photos, as well as corresponding sketches. For that, we propose to preserve the contextual similarity between instances and their nearest neighbours, modelled as a distribution over pairwise distances, from the pre-trained teacher (photo model), to that of the student (FG-SBIR). Inspired from recent literature, we introduce a novel \textit{distillation token} into our PVT-backboned student, that is dedicated towards distilling contextual similarity. 

{However fitting this token into the existing PVT-architecture is non-trivial. Unlike other vision transformers like ViT \cite{dosovitskiy2020an}, PVT employs a reshape operation \cite{wang2021pyramid}, to reshape the resultant individual patch tokens at one level back to the feature map, for input to the next level. It follows that adding a token here naively would break this operation. We thus use our distillation token only during input to the transformer layer at each level \cite{wang2021pyramid}, and set the modified token aside before the reshaping operation. A residual connection \cite{song2017deep} thereafter connects this modified token as input to the transformer layer at the next level. Engaging the distillation token at every level like ours, helps it imbibe the inductive bias modelled by the pyramidal structure of PVT, thus facilitating better distillation.}

Finally, on the back of a pilot study (\cref{sec:pilot}), we uncover a widespread problem with standard triplet loss training -- that training is highly unstable, reflected in the highly oscillating evaluation accuracies noted at every 100$^\text{th}$ training iteration. For that, we take inspiration from literature on stabilising GAN training \cite{yaz2018unusual} on Exponential Moving Average \cite{yaz2018unusual} -- a strategy that employs a moving average on model parameters iteratively, with a higher priority (mathematically exponential) to recent iterations over earlier ones. 

To sum up: (i) We propose a strong baseline for FG-SBIR, that overshoots prior arts by $\approx$10\% (ii) We achieve this by putting forward two simple designs each tackling a key problem facing the community: inadequate latent space separation, and sketch data scarcity. (iii) We introduce a simple modification to the standard triplet loss to explicitly enforces separation amongst photos/sketch instances. (iv) We devise a knowledge distillation token in PVT \cite{wang2021pyramid} that facilitates better knowledge distillation in training from unlabelled data. 
{Finally, apart from surpassing prior arts significantly, our model also shows encouraging results on generalising to new classes, without any paired sketches.}
 
\section{Related Works}

\vspace{-0.2cm}
\keypoint{Fine-grained SBIR (FG-SBIR):}
{FG-SBIR aims at retrieving one particular photo from a gallery of specific-category corresponding to a query-sketch. Initially proposed as a deep triplet-ranking based \emph{siamese network}~\cite{yu2016sketch}, FG-SBIR has progressively improved via attention-based modules with a higher order retrieval loss~\cite{song2017deep}, textual tags~\cite{song2017fine,chowdhury2023scenetrilogy}, self-supervised pre-training ~\cite{pang2020solving}, hierarchical co-attention~\cite{sain2020cross}, cross-category generalisation \cite{pang2019generalising}, and reinforcement learning~\cite{bhunia2020sketch}. Overall, they aim to learn a joint embedding space so as to reduce the cross-modal gap \cite{ha2018a, sain2020cross}, typically using a triplet ranking loss \cite{yu2016sketch} that aims to bring a sketch closer to its paired photo while distancing it from others. Some have optimised target-photo rank \cite{bhunia2020sketch}, fused classification loss with ranking \cite{huang2017deep}, used multi-task attribute loss \cite{song2016deep}, or other loss modules in self-supervised pretext tasks \cite{bhunia2021vectorization, pang2020solving} for efficiency.
Others addressed sketch-specific traits like style-diversity~\cite{sain2021stylemeup}, data-scarcity~\cite{bhunia2021more} and redundancy of sketch-strokes~\cite{bhunia2022sketching} in favor of better retrieval. Arguing against such complex frameworks, we aim for a simpler yet stronger FG-SBIR pipeline, that can generalise well across categories even in low-data regime.}

\begin{figure*}[!hbt]
    \includegraphics[width=0.49\linewidth,trim={0 1.15cm 0 2.6cm},clip]{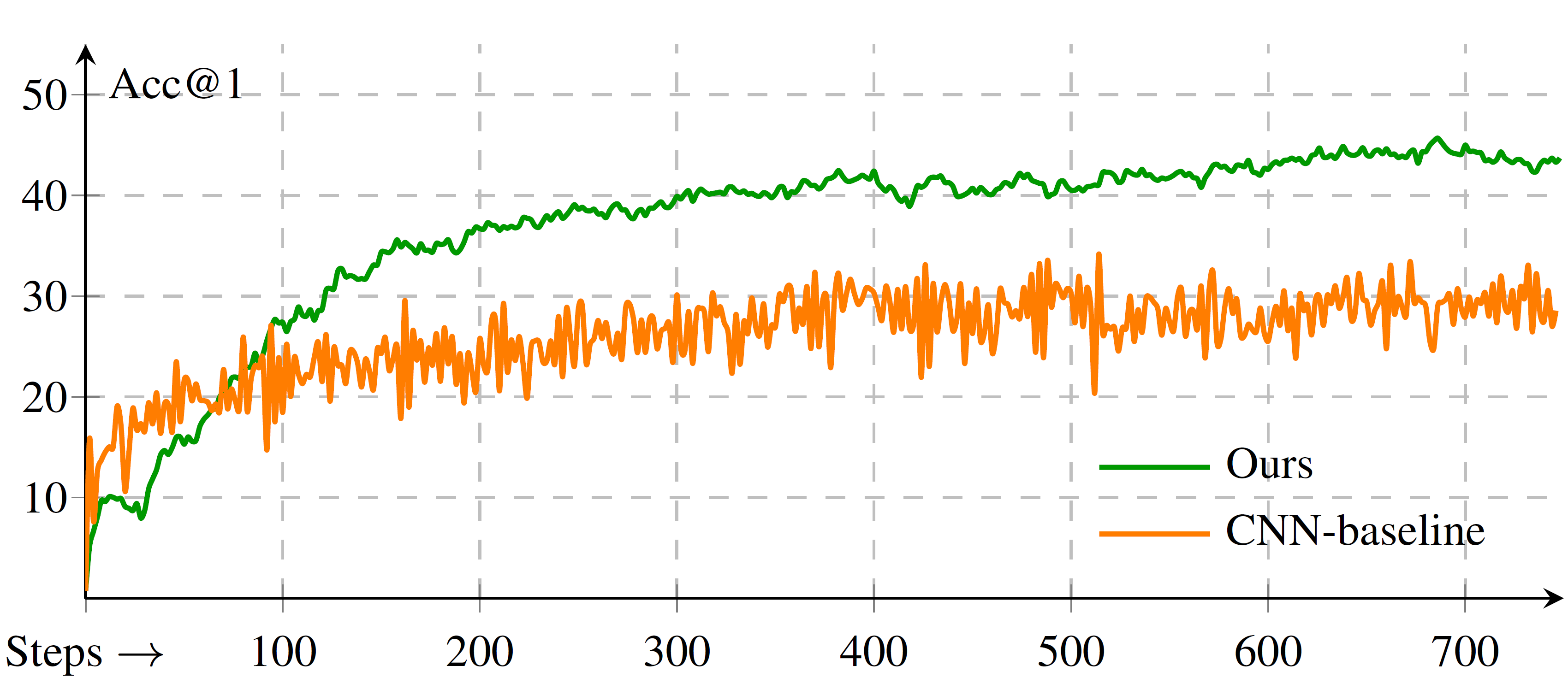}
    \includegraphics[width=0.24\linewidth]{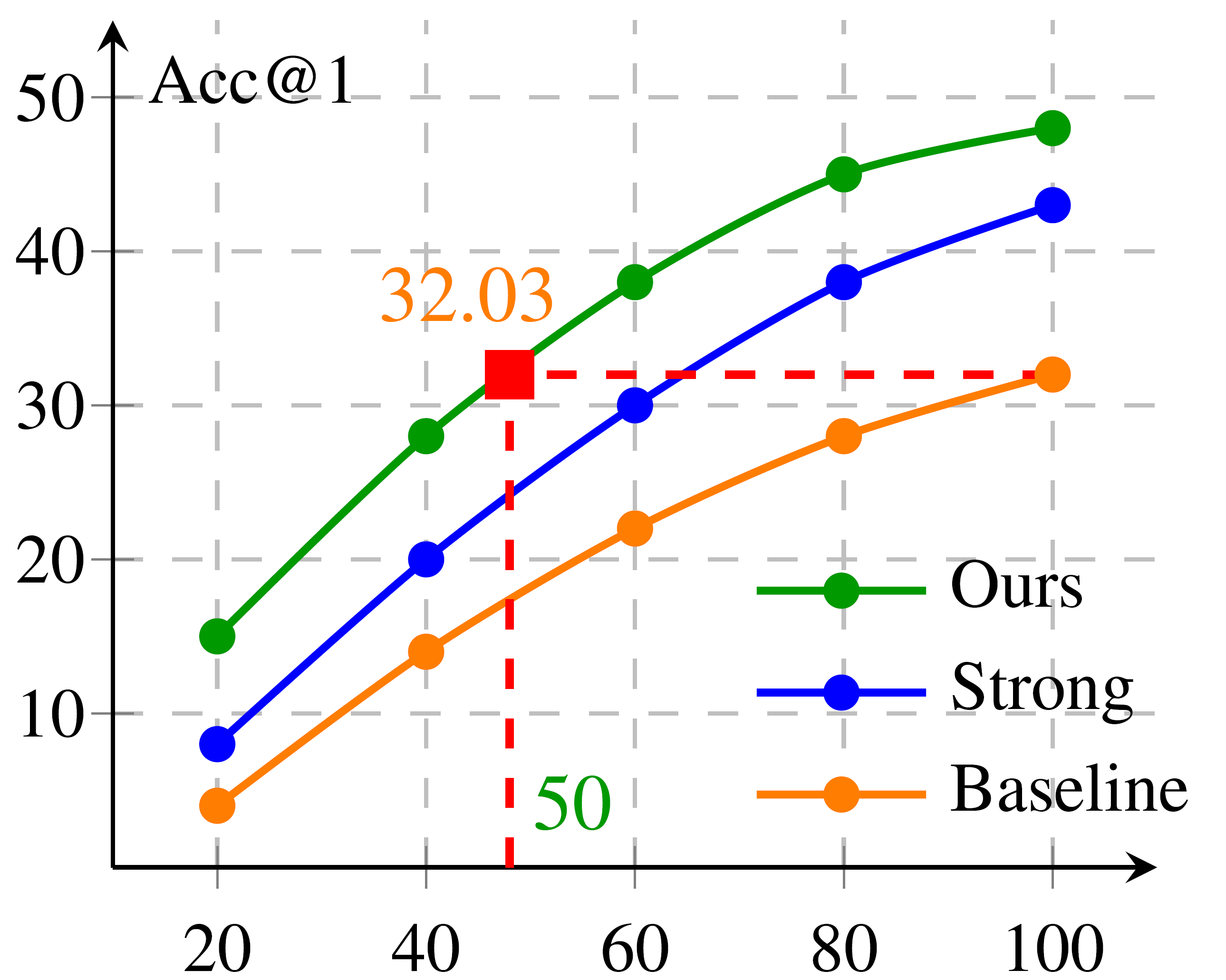}
    \includegraphics[width=0.25\linewidth]{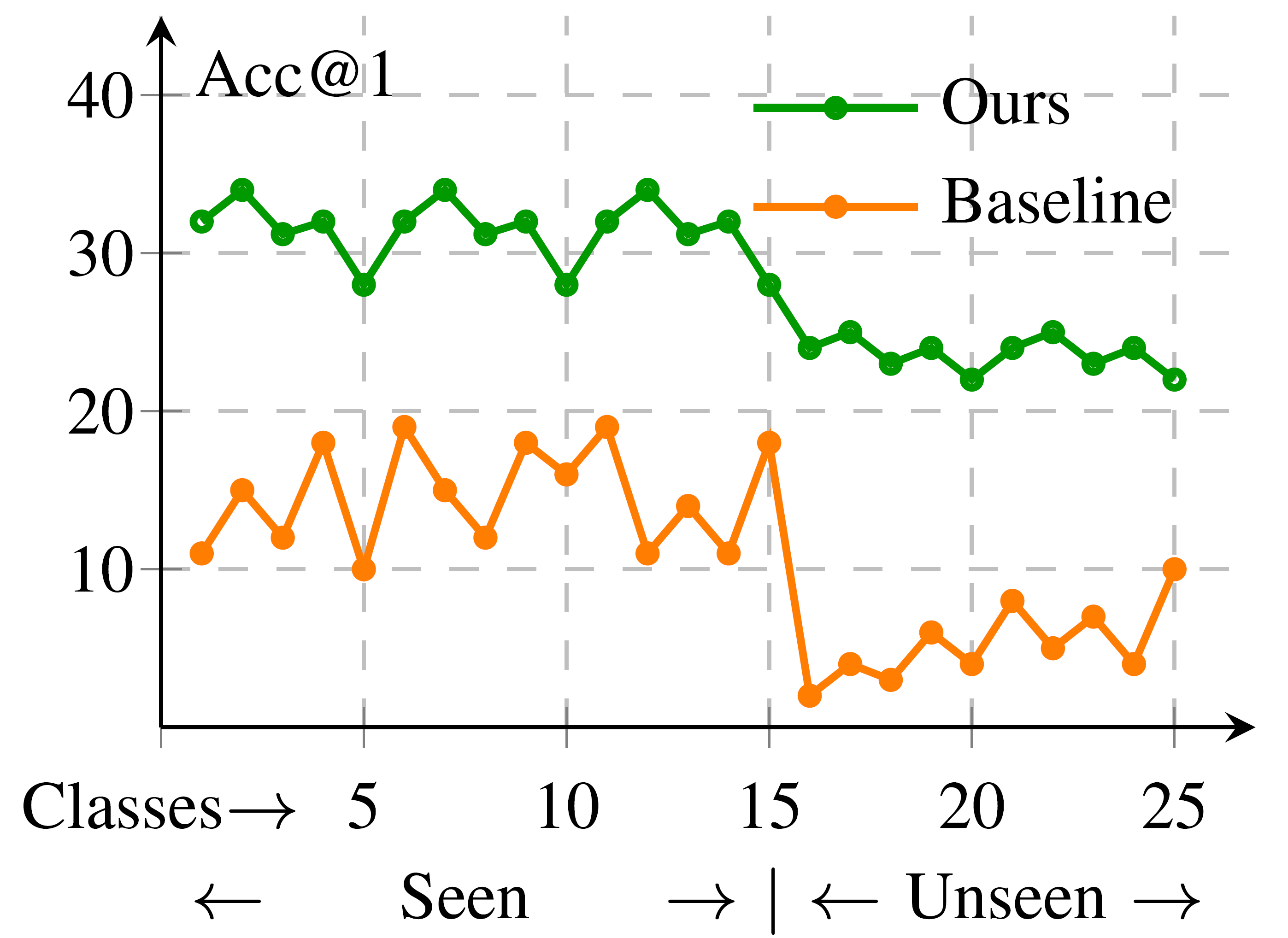}
    \vspace{-0.25cm}
    \caption{{Left: Training stability of a \orange{baseline} vs {ours}. Study on variable data-size (center) and cross-category generalisation (right).}}
	\label{fig:pilot}
	\vspace{-0.6cm}
\end{figure*}

\keypoint{Transformers in Computer Vision:} 
{Transformers \cite{vaswani2017attention} are end-to-end neural networks that leverage self-attention mechanism \cite{chaudhari2019attentive} for modelling sequential data.
Vision transformers \cite{dosovitskiy2020image, liu2021swin, wang2021pyramid} win over traditional CNNs \cite{SimonyanZ14a} in their ability to {model long-range dependencies} in sequential data (here, visual patches) thus {learning a stronger feature representation} than CNNs \cite{dosovitskiy2020image}.
Since the seminal work of ViT~\cite{dosovitskiy2020image} that introduced image patches as input to transformer layers, it has been improved further via convolution components \cite{wu2021cvt}, attention-guided knowledge distillation for data-efficiency \cite{touvron2021training} and a feature pyramid \cite{wang2021pyramid, wang2021pvtv2} modeling inductive bias. Common applications include object detection \cite{carion2020end}, image super-resolution \cite{yang2020learning}, image synthesis \cite{jiang2021transgan}, etc. Despite their recent usage in fine-grained image recognition \cite{ji2020attention, liu2021transformer}, only few have employed transformers in encoding sketches~\cite{sampaio2020sketchformer,devlin2018bert} Such transformers however have not been used for FG-SBIR. }
{In this paper, we adapt transformers for the first time in FG-SBIR to propose a strong baseline that would outperform existing state-of-the-arts and we hope research would progress further considering our baseline as a standard baseline.
}

\keypoint{Training from Unlabelled Data:}
{
Mainly two frontiers of research have emerged in taking advantage of unlabelled data --  self-supervised and semi-supervised learning.
The former spans over a large literature~\cite{jing2020selfSup}, including generative models~\cite{goodfellow2014generative} like VAEs~\cite{kingma2013auto}, contrastive learning approaches~\cite{chen2020simple, he2020momentum}, clustering \cite{caron2018deep}, etc. Furthermore, several pre-text self-supervised tasks have been explored like image colorization \cite{zhang2016colorful}, in-painting \cite{pathak2016context}, etc. Recent applications in sketch include a pre-text task of solving jigsaw puzzles \cite{pang2020solving}, and learning from the dual representative nature of sketches \cite{bhunia2021more,bhunia2021vectorization}. Semi-supervised learning aims at exploiting large unlabeled data together with sparsely labeled data to improve model performance. Common approaches include entropy minimization \cite{grandvalet2004semi}, pseudo-labeling \cite{sohn2020fixmatch}, or consistency regularization \cite{miyato2018virtual}. While \textit{pseudo-labeling} employs confident prediction \cite{sohn2020fixmatch} from trained classifiers to create \emph{artificial labels} \cite{lee2013pseudo} for unlabelled data, \textit{consistency regularisation} learns a classifier by promoting consistency in predictions between different views of unlabeled data, either via soft \cite{muller2019does} or hard \cite{sohn2020fixmatch} pseudo-labels. Recently, data scarcity in SBIR was handled, by generating more sketches for unlabelled photos \cite{bhunia2021more} in a semi-supervised setup. }

{
We focus on one of its related paradigm of knowledge distillation ~\cite{ba2014deep} that aims at transferring knowledge of a pre-trained \textit{teacher} network to a \textit{student}. While some leverage output logits~\cite{ba2014deep} others focus on hidden layers~\cite{romero2014fitnets} or  attention-maps~\cite{zagoruyko2017KDattention} of pre-trained teachers for the same. Improvising further, self-distillation~\cite{bagherinezhad2018label} employed the same network for both student and teacher models, whereas a multi-exit strategy~\cite{phuong2019distillation} optimised compute via multiple-exits at different depths for adaptive inference. Common applications include  object detection~\cite{deng2019video}, semantic segmentation~\cite{he2019semantic}, depth-estimation~\cite{pilzer2019depth}, etc.  {Contrarily, FG-SBIR demands: (a) instance level discrimination (b) cross-modal one-to-one correspondence. As collecting sketch-photo labelled pairs is costly, we train a teacher from abundant unlabelled photos. The instance-discrimination thus learned in photo space, is distilled to a student FG-SBIR model to make it more discriminative.}
}

\vspace{-0.2cm}
\section{Pilot Study: Delving deeper into FG-SBIR} \label{sec:pilot}
\vspace{-0.1cm}

\keypoint{Training stability:} 
{{
Being an instance-level cross-modal retrieval problem, training an FG-SBIR \cite{sain2020cross}
model is often found to be unstable, as evaluation accuracy oscillates significantly during training. Hence, it is evaluated frequently to capture the best accuracy. Being the first to consider this, we first analyse by plotting (\cref{fig:pilot}) test-set evaluation at every 100 training-steps of {an \emph{existing} baseline FG-SBIR model with VGG-16 backbone, trained over standard triplet loss \cite{yu2016sketch}}. Towards reducing such instability, we design a \textit{stronger} baseline model, which we hope will be a standard baseline for future research. 
}}

\keypoint{Variable dataset size:}
{
Existing literature shows FG-SBIR model performance to suffer from scarcity of training sketch-data \cite{bhunia2021more}. This leads us to wonder if unlabelled photos can be leveraged to train an FG-SBIR model to reduce {the data annotation bottleneck} on sketches for training. Consequently, we conduct a study to analyse how an FG-SBIR model performs when training data-size is varied. {Accordingly we compute and plot the performances of \textit{existing} and our \textit{stronger} baselines (\cref{sec:basemodel}) on varying dataset-size in \cref{fig:pilot}. {As both baselines, perform poorly on decreasing labelled training data, we explore if unlabelled photos can boost performance in such a scenario.} We thus design a distillation strategy from unlabelled photos, following which, our method at $\approx$50\% training data, matches the accuracy (32.03\%) of an existing baseline at 100\% training data (red square in \cref{fig:pilot}), thereby justifying our paradigm.}
}

\keypoint{Classes without any sketch-photo pairs:}
{
Besides training data scarcity, generalisation to unseen classes for multi-category FG-SBIR \cite{bhunia2022adaptive}, is also one of the challenging issues in FG-SBIR.
Considering a realistic scenario, sketch-photo pairs might not always be available for all classes, however photos pertaining to such classes can be curated fairly easily. Therefore, given a model trained on classes having sketch-photo pairs, we aim to discover how well can it perform on classes lacking paired sketches.
We thus conduct a small pilot study by taking the first 25 classes (alphabetically) from Sketchy \cite{sangkloy2016sketchy} -- 
{15 seen training (sketch-photo pairs available) and 10 testing (only photos available), in an FG-SBIR setting. 
\cref{fig:pilot} shows baseline model to perform significantly better on classes with access to sketch-photo pairs but fails poorly on classes lacking paired sketches. 
The main challenge therefore is to leverage the knowledge of classes having unlabelled photos, to preserve the model's accuracy.
In doing so, our method maintains a stable performance via its knowledge distillation paradigm.
}}

\begin{figure*}[t]
\begin{center}
  \includegraphics[width=\linewidth]{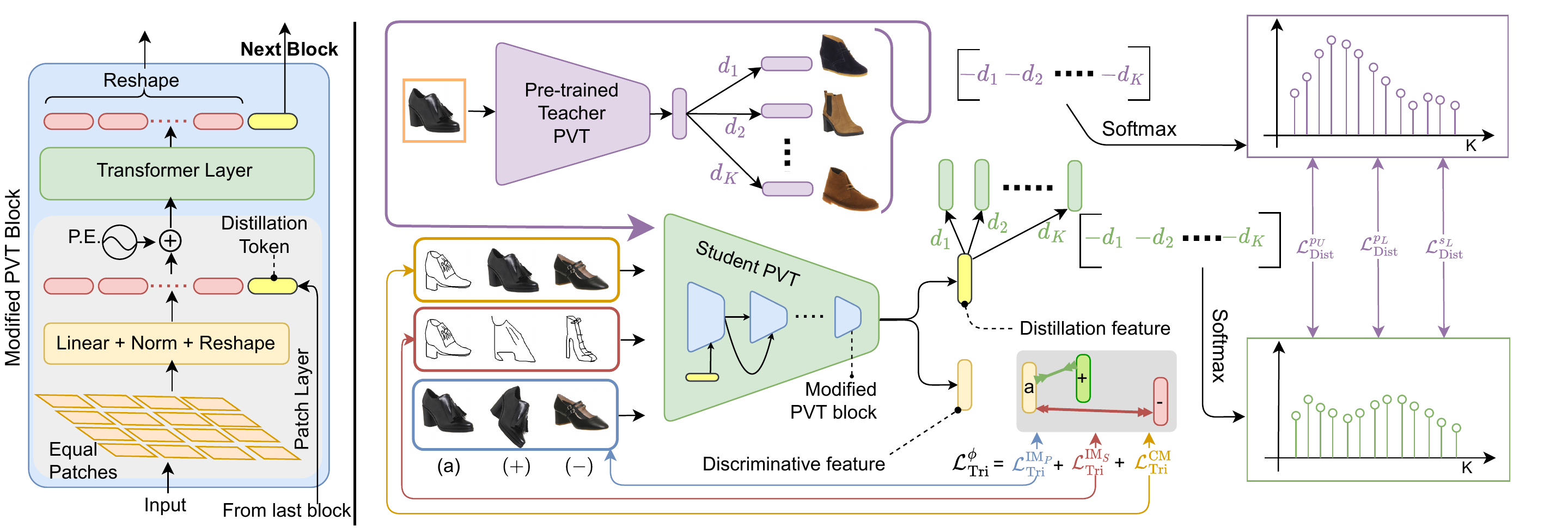}
\end{center}
\vspace{-0.7cm}
  \caption{{
    Using PVT as a backbone, a teacher pre-trained on unlabelled photos, distills the contextual similarity among features in its latent space to a student, which also learns a discriminative latent via cross- and intra-modal triplet losses. Distillation occurs via a learnable distillation token (shown as inset) introduced in the student's PVT backbone, in three ways: (i) from unlabelled photos, (ii) labelled photos and (iii) by aligning sketches to their paired photos in student-space by distilling contextual similarity of labelled photos in teacher-space. 
  }}
\vspace{-0.55cm}
\label{fig:Framework}
\end{figure*}

\vspace{-0.2cm}
\section{A Stronger FG-SBIR baseline} \label{sec:basemodel}
\vspace{-0.15cm}
\keypoint{Overview:}{Unlike CNN-based existing FG-SBIR baselines \cite{song2017deep,yu2016sketch} that learn a strong sketch-embedding function over a cross-modal triplet loss, we enhance the paradigm with three distinct modifications: (i) Employ vision transformer, particularly PVT \cite{wang2021pyramid} as our backbone after thorough analysis (\cref{sec:experiments}), as unlike CNNs having local receptive fields, vision transformers ensure a global receptive field, modelling better feature representations. (ii) Besides cross-modal triplet loss\cite{yu2016sketch}, we formulate an intra-modal triplet loss that helps in discriminative learning. (iii) Towards increasing training stability we follow GAN-literature where training GANs are often unstable~\cite{yaz2018unusual}, in employing Exponential Moving Average (EMA) to our paradigm \cite{yaz2018unusual}.
} 

\keypoint{Architecture:} In a nutshell, PVT~\cite{wang2021pyramid} generates feature maps at different scales by operating on an image $I \in \mathbb{R}^{H\times W \times 3}$ (sketch/photo) over multiple ($m$) levels. Given input $I_l \in \mathbb{R}^{H_{l-1} \times W_{l-1} \times C_{l-1}}$ at level $l \in [1,m]$: 
{\textit{(a)}} a patch-embedding layer extracts patch-wise features ($\text{p}^i \in \mathbb{R}^{\frac{H_{l-1}}{\text{p}_l} \times \frac{W_{l-1}}{\text{p}_l} \times C_{l-1}}; i \in [1, \frac{H_{l-1} W_{l-1}}{\text{p}_l^2}]$) of patch-size $\text{p}_l\times \text{p}_l$. 
{\textit{(b)}} They are passed via a transformer layer to obtain patch-tokens, which
{\textit{(c)}} are reshaped to a down-scaled feature map $F_l \in \mathbb{R}^{\frac{H_{l-1}}{\text{p}_l} \times \frac{W_{l-1}}{\text{p}_l} \times C_l}$ as input for next level ($I_{l+1}$). Following \cite{wang2021pyramid} we use $m$ = $4$ levels, keeping $\text{p}_l$ = $4$ per level, to obtain the global average pooled final feature $f_I \in \mathbb{R}^{d}$ for retrieval.

\keypoint{Cross-modal (CM) Triplet Loss:} Taking independent embedding of a sketch ($f_s$) as anchor, {traditional \cite{yu2016sketch} cross-modal triplet loss} $\mathcal{L}_{\text{Tri}}^\mathrm{CM}$ aims to minimise its distance from its paired photo embedding ($f_{p}$) while maximising that from a non-matching one ($ f_{n}$) in a joint embedding space. Using $m_\mathrm{CM}$ as the margin hyperparameter and $\delta(\cdot,\cdot)$ as a distance function where $\delta(a,b) = ||a-b||^2$, we have,

\vspace{-0.5cm}
\begin{equation}
    \begin{aligned}
    \mathcal{L}_{\text{Tri}}^\mathrm{CM} =  \max \{0, m_\mathrm{CM} + \delta(f_s, f_{p}) - \delta(f_s, f_{n})\}
    \end{aligned}
    \label{equ:CM}
    \vspace{-0.1cm}
\end{equation}

\noindent \textbf{Intra-modal (IM) Triplet Loss: } {Despite separating \textit{sketches} from non-matching photos, $\mathcal{L}_{\text{Tri}}^\mathrm{CM}$ often pushes visually similar \textit{photos} closer, resulting in sub-optimal latent space. Consequently, we focus on intra-modal feature-separation via two sets of intra-modal triplet losses. Having multiple sketches per photo, we take features of query-sketch as the anchor ($f_s$), another sketch of the same target photo ($p$) as its positive ($f_{s^+}$), and that of a random non-matching photo as the negative ($f_{s^-}$). 
Given an FG-SBIR paradigm, as no separately paired positive photo exists for any photo ($p$), we curate a structural augmentation of $p$ as $f_{p^t}$ = \texttt{Perspective}(\texttt{Rotate}($p$)) where \texttt{Rotate}($\cdot$) randomly rotates an image (-45\textdegree to +45\textdegree), and \texttt{Perspective}($\cdot$) introduces random perspective transformation on it. Bearing semblance with a sketch, this only affects the structure of an image, instead of its colour/texture. Taking $p$ as anchor, $p_T$ as its positive and a random negative ($n$) we compute intra-modal losses, with $m_\mathrm{IM}^p$ and $m_\mathrm{IM}^s$ as respective margin values for photo and sketch modalities.
\vspace{-0.1cm}         
\begin{equation}
    \begin{aligned}
    &\mathcal{L}_{\text{Tri}}^{\mathrm{IM}_p} =  \max \{0, m_\mathrm{IM}^p + \delta(f_{p}, f_{p^t}) - \delta(f_{p}, f_{n})\}, \\
    &\mathcal{L}_{\text{Tri}}^{\mathrm{IM}_s} =  \max \{0, m_\mathrm{IM}^s + \delta(f_s, f_{s^+}) - \delta(f_s, f_{s^-})\}
    \end{aligned}
    \label{equ:intramodal}
    \vspace{-0.1cm} 
\end{equation}
}

\vspace{+0.0cm}
\noindent \textbf{Exponential Moving Average:} Following literature on instability of GAN-training \cite{yaz2018unusual}, or teacher-training in semi-supervised contrastive learning \cite{cai2021exponential}, we employ EMA \cite{cai2021exponential} to increase training stability. While uniform average of parameters stabilises training \cite{yaz2018unusual}, it gets limited by memory. A moving average (MA), with exponential priority (EMA) to recent iterations over earlier ones is thus preferred as,
\vspace{-0.2cm}
{
\begin{equation}
    \begin{aligned}
    \theta^t_\text{EMA} = \beta\theta^{t-1}_\text{EMA} + (1-\beta)\theta^{t}
    \end{aligned}
\end{equation}
where $\theta^0_\text{EMA}=\theta^0$, and $\beta$ is a hyperparameter deciding the rate at which early iterations fade.
With $\lambda_{1,2}$ as weighting hyperparameters, we have our overall training loss:
\vspace{-0.1cm} 
\begin{equation}
    \begin{aligned}
    \mathcal{L}_{Trn} = \mathcal{L}_{\text{Tri}}^\mathrm{CM} 
                + \lambda_1 \mathcal{L}_{\text{Tri}}^{\mathrm{IM}_{p}}
                + \lambda_2 \mathcal{L}_{\text{Tri}}^{\mathrm{IM}_{s}} 
    \end{aligned}
    \label{equ:stronger_loss}
    \vspace{-0.1cm} 
\end{equation}
}

\vspace{-0.4cm}
\section{Training with Unlabelled Photos via KD} \label{sec:distillation}
\vspace{-0.20cm}

Once we have a stronger baseline, we aim to harness the learning potential of unlabeled photos, towards training a strong FG-SBIR model \cite{bhunia2021more}. We thus employ a knowledge distillation paradigm~\cite{hinton2015distilling}, that transfers the photo instance discrimination potential from a pre-trained `teacher' ($\Omega$) trained on unlabelled photos to our `student' ($\phi$) FG-SBIR model for cross-modal retrieval. Given a limited amount of labelled sketch-photo pairs ${\mathcal{D}_{L} = \{ ({p_{L}^{i}}; {s_L^{i}})\}_{i}^{N_{L}}}$ for our student FG-SBIR model, and a larger set of unlabelled photos ${\mathcal{D}_{U} = \{ {p_{U}^{i}} \}_{i}^{N_{U}}}$, for pre-training our teacher ($N_{U} \gg N_L$), our goal is to improve retrieval accuracy of our student FG-SBIR model, using both $\mathcal{D}_{L}$ and  $\mathcal{D}_{U}$ (no paired sketches).


However applying this KD paradigm is non-trivial to our setting. 
While conventional KD usually transfers knowledge via logit-distillation \cite{hinton2015distilling} for classification problem, ours is a cross-modal retrieval setup where the output is a continuous $d$-dimensional feature in a joint-embedding space. Moreover, if one naively regresses between features from a teacher and student for sketch and photo branches, it might suffer from incompatibility as the two embedding spaces are different. Also, if teacher \cite{ba2014deep} and student embedding dimensions are different, an additional feature transformation layer \cite{romero2014fitnets} is needed to match them. While distilling through pair-wise distance~\cite{saputra2019distilling, bhunia2021more} might be an option, it fails to transfer the structural knowledge of the teacher's entire latent space to the student, as it focuses on only one pair at a time.  We thus aim to design a distillation framework that considers the contextual similarity amongst samples in the teacher's embedding space and preserves its structural knowledge while distilling to the student.

\keypoint{Modified PVT for Distillation:}
Concretely, our network involves two models -- a teacher $\Omega(\cdot)$ pre-trained large-scale photos, and a learnable FG-SBIR student $\phi(\cdot)$, both of which uses PVT \cite{wang2021pyramid} as backbone feature extractor. While the teacher's backbone PVT remains unchanged, we follow recent transformer literature \cite{touvron2021training} to introduce a learnable distillation token `$\Delta$' $\in \mathbb{R}^d$, that allows our model to learn from the output of the teacher, while remaining complementary to the feature extracted from PVT \cite{wang2021pyramid}. However, naively concatenating a token is infeasible, as unlike other vision transformers \cite{touvron2021training}, PVT involves reshaping (for down-scaling) a fixed number of individual patch tokens to the subsequent feature map, which would be imbalanced on adding one extra token. Therefore, we focus on the second step  of PVT-block \cite{wang2021pyramid} involving transformer layer (\cref{sec:basemodel}), which can accommodate a variable number of tokens. Accordingly, at every level, `$\Delta$' is fed to the transformer layer concatenated with rest of the $N$ image patch tokens ($\{\text{p}_i\}^N_{i=1} \in \mathbb{R}^{N\times d}$), to obtain the resultant set of tokens as $\{\text{p}_i\}^{N+1}_{i=1} \in \mathbb{R}^{(N+1)\times d}$. Before reshaping, $\Delta$ is excluded to prevent dimensional mismatch, and fed again similarly to the transformer layer at the next level via a residual connection (\cref{fig:Framework}). 
Being processed at every level, `$\Delta$' not only accommodates the output knowledge of the teacher network but also, imbibes the inductive bias \cite{wang2021pyramid} contributed by the pyramidal structure of PVT. From the final layer, the student outputs two features: a \textit{discriminative} ($f \in \mathbb{R}^d$), and a \textit{distillation} feature ($\mu \in \mathbb{R}^d$).

\keypoint{Pre-training Teacher:} 
The teacher $\Omega(\cdot)$, is trained on photos of both labelled and unlabelled sets $\mathcal{G} = \mathcal{D}_{U} \cup \mathcal{D}_L^p$, over an intra-modal triplet loss following Eqn. \ref{equ:intramodal}.

\keypoint{Training FG-SBIR Student:} 
The student's \textit{discriminative} feature is used to train over a combination ($\mathcal{L}^L_{Trn}$) of cross-modal and intra-modal triplet losses on labelled data ($D_L$) following Eqn. \ref{equ:stronger_loss}.
Intra-modal triplet loss over unlabelled photos is leveraged to harness the potential of unlabelled photos $\mathcal{D}_U$ with a weighting hyperparameter $\lambda_3$ as, 
\vspace{-0.2cm}
\begin{equation}
    \mathcal{L}_{\text{Disc}}^\phi = \mathcal{L}^L_{Trn} + \lambda_3 \mathcal{L}_{\text{Tri}}^U
    \label{equ:Triplet_student}
    \vspace{-0.2cm}
\end{equation}

In order to transfer \cite{romero2014fitnets} the instance-discriminative knowledge of photo domain from the pre-trained teacher $\Omega(\cdot)$ to improve the cross-modal instance level retrieval of student FG-SBIR model $\phi(\cdot)$, we leverage both unlabelled photos and sketch/photo pairs from labelled data during knowledge distillation. First, we pre-compute the features of all the unlabelled photos $\mathcal{D}_U$ using the frozen teacher model as $\mathbf{F}^\Omega_{U} = \{f_{p^{i}}^\Omega\}_{i=1}^{N_U} \text{where} \; f_{p^i}^\Omega = \Omega(p^i_U) \in \mathbb{R}^d$.

Now a very naive way of distillation \cite{hinton2015distilling} would be to pass $p_U^i$ through student model $\phi(\cdot)$ to obtain the distillation feature $\mu_{p^i}^\phi = \phi(p_U^i)$, and directly regress it against $f_{p^i}^\Omega$ considering it as ground-truth.
Alternatively, one may regress the feature-wise distance between two photos ($p^i_U, p^j_U$) in a similar fashion from teacher to student model \cite{hinton2015distilling}. However, we focus on preserving the structure of the embedding space of the teacher while distillation to the student. We thus calculate the pairwise distance of $f_{p^{i}}^\Omega$ from its $K$ nearest neighbours $\{f_{p^{r_1}}^\Omega, \cdots, f_{p^{r_K}}^\Omega\}$, as $\mathbf{D}_{p^i}^\Omega =\{ \delta(f_{p^{i}}^\Omega, f_{p^{r_j}}^\Omega)\}_{j=1}^K$. Equivalently, we pass $p_U^i$ and its K-nearest neighbours $\{p^{r_1}, \cdots, p^{r_K}\}$ via the student model to obtain corresponding distillation features $\mu_{p^{i}}^\phi$ and $\{\mu_{p^{r_1}}^\phi, \cdots, \mu_{p^{r_K}}^\phi\}$ respectively, thus calculating $\mathbf{D}_{p^i}^\phi =\{ \delta(\mu_{p^{i}}^\phi, \mu_{p^{r_j}}^\phi)\}_{j=1}^K$ similarly. Although one may calculate a regression loss between \cite{park2019relational} $\mathbf{D}_{p^i}^\Omega$ and  $\mathbf{D}_{p^i}^\phi$, for better stability we model them as probability distribution of pairwise-similarity amongst K-nearest neighbours in the teacher's embedding space. As pairwise similarity is negative of pairwise distances, we calculate the temperature ($\tau$) normalised softmax ($\mathcal{S}$) \cite{sain2020cross} probability as $\mathcal{S}_\tau(-\mathbf{D}_{p^i}^\Omega)$ where, 
\vspace{-0.3cm}
\begin{equation}
    \mathcal{S}_\tau(-\mathbf{D}_{p^i}^\Omega)_{r^j} = \frac{\text{exp}(-\delta(f^\Omega_{p^i},f^\Omega_{p^{r_j}})/\tau)}{\sum^{r_K}_{k=1} \text{exp}(-\delta(f^\Omega_{p^{i}},f^\Omega_{p^{r_k}})/\tau)}
\vspace{-0.2cm}
\end{equation}
Similarly obtained $\mathcal{S}_\tau(-\mathbf{D}_{p^i}^\phi)$, and $\mathcal{S}_\tau(-\mathbf{D}_{p^i}^\Omega)$ represent the structural knowledge of embedding spaces of student $\phi(\cdot)$ and teacher $\Omega(\cdot)$ respectively. The consistency constraint can therefore be defined as the Kullback-Leibler (KL) divergence \cite{matthews2016sparse} between them as,
\vspace{-0.2cm}
\begin{equation}
    \mathcal{L}_\text{KL}^{p_U} = KL(\mathcal{S}_\tau(-\mathbf{D}_{p^i}^\Omega) \; \;|| \; \; \mathcal{S}_\tau(-\mathbf{D}_{p^i}^\phi))
    \label{equ:KL-main}
    \vspace{-0.2cm}
\end{equation}
On the other hand, labelled dataset comes with sketch and photo, where, using the photos we can calculate $\mathcal{L}_\text{KL}^{p_L}$ in a similar fashion. Given the cross-modal nature of student's embedding space \cite{yu2016sketch}, we also need to align the sketches with their paired photos while preserving the existent contextual similarity of those photos with their neighbours. Being trained on photos alone, extracting sketch features via the teacher $\Omega(\cdot)$ to extract sketch-features, would be a flawed design choice. 
Considering a sketch-photo pair ($s_i,p_i$), we first obtain $\mathbf{D}_{p^i}^\Omega$ using $p_i$. However this time we use $s_i$ to calculate $\mathbf{D}_{s^i}^\phi =\{ \delta(\mu_{s^{i}}^\phi, \mu_{p^{r_j}}^\phi)\}_{j=1}^K$. 
Now, calculating $\mathcal{L}_\text{KL}^{s_L}$ between these two resultant probabilities implies maintaining the same contextual similarity for sketches as that for their paired photos, as guided by the teacher model. 
With $\lambda_{4,5}$ as respective weighting hyperparameters, the total distillation loss becomes,
\vspace{-0.2cm}
\begin{equation}
    \mathcal{L}^\phi_\text{Dist} = \mathcal{L}_\text{KL}^{p_L} 
                                + \lambda_4 \mathcal{L}_\text{KL}^{s_L}
                                + \lambda_5 \mathcal{L}_\text{KL}^{p_U}
    \label{equ:Distillation_student}
    \vspace{-0.25cm}
\end{equation}
\noindent Summing up, our student model is trained from a weighted (hyperparameter $\lambda_6$) combination of two losses as:
\vspace{-0.2cm}
\begin{equation}
    \mathcal{L}^\phi_{trn} = \mathcal{L}^\phi_\text{Disc} + \lambda_6 \mathcal{L}^\phi_\text{Dist}
    \label{equ:student}
    \vspace{-0.20cm}
\end{equation}


\section{Experiments}\label{sec:experiments}
\vspace{-0.2cm}
\keypoint{Datasets:} {We use two publicly available datasets, QMUL-Chair-V2 and QMUL-Shoe-V2 \cite{yu2016sketch, sain2020cross}. They contain 2000 (400) and 6730 (2000) sketches (photos) respectively with fine-grained sketch-photo associations. We keep 1275 (300) and 6051 (1800) sketches (photos) from QMUL-Chair-V2 and QMUL-Shoe-V2 respectively for training while the rest is used for testing. We also use Sketchy \cite{sangkloy2016sketchy} which contains 125 categories with 100 photos each, having at least 5 sketches per photo with fine-grained associations. While, training uses a standard (90:10) train-test split \cite{sangkloy2016sketchy}, during inference we construct a challenging gallery using photos across one category for retrieval. Besides such labelled training data, we use all 50,025 photos of UT-Zap50K \cite{yu2014fine} and 7,800 photos \cite{pang2020solving} collected from shopping websites, including IKEA, Amazon and Taobao, as unlabelled photos for shoe and chair retrieval, respectively. For Sketchy we use its extended version with 60,502 additional photos \cite{liu2017deep} introduced later for training from unlabelled data.
}

\keypoint{Implementation Details:}{
ImageNet \cite{deng2009imagenet} pre-trained PVT-\textit{Large} \cite{wang2021pyramid} model extracts features from $224$ $\times$ $224$ resized images, keeping patch-size ($\text{p}_l$) of 4 at each level and 1, 2, 5, and 8 spatial-reduction-attention heads \cite{wang2021pyramid} in 4 successive levels, with the final feature ($\mu$ and $f$) having size 512. Implemented via PyTorch \cite{paszke2017automatic}, our model is trained using Adam-W optimiser\cite{loshchilov2019decoupled} with momentum of 0.9 and weight decay of 5e-2, batch size of 16, for 200 epochs, on a 11 GB Nvidia RTX 2080-Ti GPU. Initial learning rate is set to 1e-3 and decreased as per cosine scheduling \cite{loshchilov2017sgdr}. Determined empirically, $m_\mathrm{CM}$, $m_\mathrm{IM}^s$, $m_\mathrm{IM}^p$ and $\tau$ are set to 0.5, 0.2, 0.3 and 0.01, while $\lambda_{1 \rightarrow 6}$ to 0.8, 0.2, 0.4, 0.4, 0.7  and 0.5 respectively. Following \cite{sain2020cross} we use Acc.@q, i.e. percentage of sketches having true matched photo in the top-q list.
}

\vspace{-0.3cm}
\subsection{Competitors} \label{sec:competitors}
\vspace{-0.20cm}
\noindent {
We compare against:
\textbf{(i) State-of-the-arts} (\textbf{SOTA}): 
\textit{Triplet-SN}~\cite{yu2016sketch} trains a Siamese network on cross-modal triplet loss to learn a discriminative joint sketch-photo embedding space. 
While \textit{HOLEF-SN}~\cite{song2017deep} uses a spatial attention module over Sketch-a-Net \cite{yu2016sketchAnet} backbone,
\textit{Jigsaw-SN}~\cite{pang2020solving} employs jigsaw-solving pre-training over mixed patches of photos and edge-maps followed by triplet-based fine-tuning for better retrieval.
\textit{Triplet-RL}~\cite{bhunia2020sketch} leverages triplet-loss based pre-training, followed by RL based fine-tuning for \textit{on-the-fly} retrieval. We report its results only on completed sketches as early retrieval is not our goal. \textit{StyleVAE} \cite{sain2021stylemeup} meta-learns a VAE-based disentanglement module for a style-agnostic retrieval.
Following \cite{bhunia2021more} \textit{Semi-sup-SN} trains a sequential photo-to-sketch generation model that outputs pseudo sketches as labels for unlabelled photos, to semi-supervise retrieval better.
\vspace{-0.32cm}
{
\setlength{\tabcolsep}{5pt}
\begin{table}[H]
    \centering
    \renewcommand{\arraystretch}{0.85}
    \footnotesize
    \caption{\footnotesize{Quantitative comparison of pipelines.}}
    \vspace{-0.25cm}
    \label{tab:quantitative_main}
    \begin{tabular}{cclcccc} 
    \toprule
    \multicolumn{3}{c}{\multirow{2}{*}{Methods}} &
    \multicolumn{2}{c}{Chair-V2 (\%)} &
    \multicolumn{2}{c}{Shoe-V2 (\%)}
    \\ \cmidrule(lr){4-5}\cmidrule(lr){6-7} 
    & & & Top-1 & Top-10 & Top-1 & Top-10 
    \\ \cmidrule(lr){1-5}\cmidrule(lr){6-7} 
    \multicolumn{2}{c}{\multirow{6}{*}{\rotatebox[origin=c]{90}{SOTA}}}
    & Triplet-SN~\cite{yu2016sketch}        & 47.45  & 84.32  & 28.71  & 71.56  \\
    & & HOLEF-SN~\cite{song2017deep}        & 50.41  & 86.31  & 31.24  & 74.61  \\
    & & Jigsaw-SN~\cite{pang2020solving}    & 53.41  & 87.56  & 33.51  & 76.86  \\
    & & OnTheFly \cite{bhunia2020sketch}    & 54.54  & 88.61  & 34.10  & 78.82  \\
    & & StyleMeUp \cite{sain2021stylemeup}  & 59.86  & 89.64  & 36.47  & 81.83  \\
    & & Semi-sup-SN \cite{bhunia2021more}   & 60.20  & 90.81  & 39.12  & 85.21  \\
    \cmidrule(lr){1-5}\cmidrule(lr){6-7}
    \multirow{17}{*}{\rotatebox[origin=c]{90}{Stronger Baseline}}
    & \multirow{6}{*}{\rotatebox[origin=c]{90}{SOTA++}}
    & Triplet-SN-ours       & 53.48  & 87.91  & 33.78  & 76.84 \\
    & & HOLEF-SN-ours       & 55.23  & 88.61  & 35.41  & 78.85 \\
    & & Jigsaw-SN-ours      & 58.51  & 88.78  & 37.64  & 79.78 \\
    & & OnTheFly-ours       & 59.18  & 89.35  & 38.62  & 81.97 \\
    & & StyleMeUp-ours      & 65.85  & 90.84  & 40.42  & 82.94 \\
    & & Semi-sup-SN-ours    & 66.86  & 91.12  & 44.35  & 86.83 \\
    \cmidrule(lr){3-5}\cmidrule(lr){6-7}
    & \multirow{10}{*}{\rotatebox[origin=c]{90}{Backbone Variants}} 
    & B-ResNet-18       & 48.42  & 85.62  & 26.61 & 70.31  \\
    & & B-ResNet-50     & 47.78  & 82.34  & 28.12 & 70.84  \\
    & & B-InceptionV3   & 55.41  & 88.21  & 34.24 & 78.56  \\
    & & B-VGG-16        & 58.23  & 88.78  & 35.85 & 80.92  \\
    & & B-VGG-19        & 61.46  & 89.16  & 37.28 & 81.01  \\
    & & B-ViT           & 38.71  & 72.65  & 16.28 & 53.42  \\
    & & B-DeIT          & 56.25  & 87.72  & 35.62 & 79.05  \\
    & & B-SWIN          & 66.34  & 91.03  & 40.71 & 82.57  \\
    & & B-CvT           & 68.42  & 91.21  & 41.58 & 83.14  \\
    & & B-CoAtNet       & 69.68  & 91.78  & 42.63 & 83.20\\
    \cmidrule(lr){3-5}\cmidrule(lr){6-7}
    & & \bf {Ours-Strong} &\bf71.22  &\bf92.18  &\bf44.18  &\bf84.68  \\
    \cmidrule(lr){1-5}\cmidrule(lr){6-7}
    \multicolumn{2}{c}{\multirow{6}{*}{\rotatebox[origin=c]{90}{Unlabelled}}} 
    & B-Edge-Pretrain       & 71.58  & 90.78  & 44.62  & 84.85\\
    & & B-Edge2Sketch       & 72.16  & 91.01  & 45.18  & 84.92\\
    & & B-Regress           & 72.65  & 91.32  & 45.45  & 85.01  \\
    & & B-RKD               & 73.02  & 91.78  & 46.18  & 85.12  \\
    & & B-PKT               & 73.45  & 91.89  & 46.66  & 85.47  \\
    & & \bf Ours-Full       & \bf 74.68  & \bf 92.79  & \bf 48.35  & \bf 85.62  \\
    \bottomrule
    \end{tabular}
    \vspace{-0.35cm}
\end{table}
}
\noindent \textbf{(ii) SOTA-Augmentation} (\textbf{SOTA++}:) 
To judge how generic our paradigm is compared to existing frameworks, we augment the mentioned state-of-the-arts by introducing our intra-modal triplet objective (\cref{sec:basemodel}) with Exponential Moving Average in their respective training paradigms.
\\\textbf{(iii) Architectural Variants:} Using popular CNN architectures like InceptionV3\cite{SzegedyVISW16}, ResNet-18,50 \cite{HeZRS16} and VGG-16,19 \cite{SimonyanZ14a} as backbone feature extractors, we explore their potential in FG-SBIR  against our strong baseline. Similarly we also explore a few existing vision transformers as backbone feature extractors for our training paradigm, namely,
\textit{B-ViT}~\cite{dosovitskiy2020image} (ViT-B16 variant),
\textit{B-DeiT}~\cite{touvron2021training} (DEiT-B variant),
\textit{B-SWIN}~\cite{liu2021swin} (SWIN-B variant),
\textit{B-CvT}~\cite{wu2021cvt} (CvT21 variant), and
\textit{B-CoAtNet}~\cite{dai2021coatnet}(CoAtNet3 with 384-dim).
\\\textbf{(iv) On training from unlabelled data: } 
Following \cite{bhunia2021more} we curate a few more baselines that could be used to leverage unlabelled photos for training. Keeping rest of the paradigm same, \textit{B-Edge-Pretrain} \cite{radenovic2018deep} naively uses edge-maps of unlabelled photos to pre-train the retrieval model. Although edge-maps bear little resemblance to sketches, we similarly design a baseline \textit{B-Edge2Sketch} that follows \cite{riaz2018learning} in converting edge-maps of photos to pseudo-sketches as labels to harness potential of unlabelled photos. As there are no previous works employing KD for FG-SBIR, we curate a few baselines offering alternatives on the paradigm of KD. These methods although repurposed for a cross-modal retrieval setting, use our network architecture ($\Omega$,$\phi$) during training. \textit{B-Regress} - directly regresses between features computed by teacher and student for both sketches and images, after matching corresponding feature dimensions via additional feature transformation layer, over a $l_2$ regression loss. \textit{B-RKD} follows \cite{park2019relational} to distill the knowledge of relative pairwise feature-distances of unlabelled photos from teacher's embedding space to that of the student, over a distance-wise distillation loss. 
{
Following \cite{passalis2018learning} off-the-shelf, for unlabelled photos, \textit{B-PKT} computes the conditional probability density of any two points in teacher's embedding space~\cite{passalis2018learning}, which models the probability of any two samples being close together. Taking `N' such samples, it obtains a probability distribution over pairwise interactions in that space. Obtaining a similar distribution over the \textit{same} samples in student's embedding space, it minimises their divergence over a KL-divergence loss. 
}
}

\vspace{-0.25cm}
\subsection{Performance Analysis}
\vspace{-0.2cm} 
\noindent\textbf{FG-SBIR pipelines:} 
{ 
\Cref{tab:quantitative_main,tab:quantitative_sketchy} compare our methods against state-of-the-arts and curated baselines.
\textit{Triplet-SN}\cite{yu2016sketch} and \textit{HOLEF-SN}\cite{song2017deep} score low due to weaker backbones of Sketch-A-Net \cite{yu2016sketchAnet}. Jigsaw-SN \cite{pang2019generalising} on the other hand improves performance, owing to self-supervised Mixed-modal jigsaw solving strategy learning structural information better. Enhanced by its RL-optimised reward function \textit{OnTheFly}\cite{bhunia2020sketch} surpasses them but fails to exceed \textit{StyleMeUp} \cite{sain2021stylemeup} (2.37\%$\uparrow$ top1 on ShoeV2), thanks to its complex meta-learned disentanglement module addressing style-diversity. 
Unlike others that are restricted to paired training data, Semi-sup-SN\cite{bhunia2021more} harnesses knowledge of unlabelled data with its additionally generated pseudo-sketch labels in achieving comparatively higher performance. However, being dependent on the usually unreliable quality of generated sketch and the instability of reinforcement learning involved, it lags behind our relatively simpler yet robust knowledge distillation paradigm, boosted with our better transformer-based feature extractor. Importantly, when augmenting our `strong-baseline' paradigm to existing SOTAs, we observe a relative rise in top-1 performance of all methods by $\approx \mathbf{4-6\%}$ overall in SOTA++ section (\Cref{tab:quantitative_main}). Despite costing a minimal memory overhead (training only), these objectives reward a considerably high accuracy boost which verifies our method to be an easy-fit and quite generic to serve as a strong FG-SBIR baseline.
}

\vspace{-0.3cm}
{\setlength{\tabcolsep}{2.5pt}
\renewcommand{\arraystretch}{0.6}
\begin{table}[!hbt]
    \centering
    \footnotesize 
    \caption{\footnotesize{Quantitative comparison of pipelines on Sketchy \cite{sangkloy2016sketchy}. }}
    \vspace{-0.24cm}
    \label{tab:quantitative_main2}
    \begin{tabular}{lcc|lcc} 
    \toprule
    \multirow{2}{*}{Methods} & \multicolumn{2}{c|}{Sketchy (\%)} &
    \multirow{2}{*}{Methods} & \multicolumn{2}{c}{Sketchy (\%)} 
    \\ \cmidrule(lr){2-3}  \cmidrule(lr){5-6}
    & Top-1 & Top-5 & & Top-1 & Top-5 \\
    \cmidrule(lr){1-3}\cmidrule(lr){4-6}
    Triplet-SN~\cite{yu2016sketch}      & 15.32  & 34.15  & B-InceptionV3  & 28.71  & 71.56  \\
    HOLEF-SN~\cite{song2017deep}        & 16.71  & 35.92  & B-VGG-16 & 18.84  & 38.63  \\
    Jigsaw-SN~\cite{pang2020solving}    & 16.74  & 36.37  & B-ViT & 7.63  & 11.23  \\
    OnTheFly \cite{bhunia2020sketch}    & 04.76  & 07.81  & B-SWIN & 32.14  & 57.68  \\
    StyleMeUp \cite{sain2021stylemeup}  & 19.62  & 39.72  & {B-CoAtNet}  & 33.63  & 59.31  \\
    \cmidrule(lr){1-3} \cmidrule(lr){4-6}
    Triplet-SN-ours     & 19.48  & 37.91  & B-Edge-Pretrain & 34.98  & 61.32 \\
    HOLEF-SN-ours       & 20.23  & 38.61  & B-Edge2Sketch & 35.81  & 61.74 \\
    Jigsaw-SN-ours      & 21.45  & 39.56  & B-Regress & 36.33  & 62.31 \\
    OnTheFly-ours       & 07.28  & 12.14  & B-RKD & 37.02  & 63.02 \\
    StyleMeUp-ours      & 22.95  & 45.84  & B-PKT & 38.62  & 63.94 \\
    \cmidrule(lr){1-3}\cmidrule(lr){4-6}
    \bf Ours-Strong     & 34.72  & 65.10  & \bf Ours-Full & 38.54  & 71.52  \\
    \bottomrule
    \end{tabular}
    \vspace{-0.4cm}
    \label{tab:quantitative_sketchy}
\end{table}
}

\keypoint{Backbone architectures:} Comparing efficiency of CNNs as backbone feature extractors (\textbf{B-CNNs}) for \textit{FG-SBIR}, we find \textit{B-VGG19} to perform best, being slightly better than \textit{B-VGG16} (by 3.23\%) at an extra memory cost (21mb), and much better than \textit{B-ResNet18} (by 10.67\%) mainly due to the latter's thinner conv-layers in Top-1 accuracy on ShoeV2. Among transformers, \textit{B-CoAtNet, B-CvT} and \textit{B-SWIN} perform much better than \textit{B-ViT} or \textit{B-DeIT} thanks to their heirarchical design and mechanisms like shifting window or convolutional tokens capturing fine-grained sparse details of a sketch via local patch-level interaction, thus representing sketches better unlike other two.
Surpassing all others, with its unique pyramidal structure imbibing inductive bias, PVT \cite{wang2021pyramid} (\textit{Ours-Strong,Full}) fits optimally to our method.

\keypoint{{Training from Unlabelled Data:}} {
While \emph{B-Edge-Pretrain} offers little gain over our stronger baseline (\textit{Ours-Strong}), augmenting edge-maps with \emph{B-Edge2Sketch} by a selected subset of strokes, to imbibe the abstractness of sketch, increases accuracy reasonably.
Lower scores of \textit{B-Regress} is largely due to its metric-based regression across two different embedding spaces causing misalignment, whereas attending to one pair at a time, \textit{B-RKD} \cite{park2019relational} fails to preserve the structural knowledge of the embedding space, thus scoring below \textit{B-PKT} (by 0.48 top1 on ShoeV2).
Although \textit{B-PKT} \cite{passalis2018learning} is slightly similar to our paradigm of preserving the latent space, lacking cross-modal discrimination objective \cite{yu2016sketch} and therefore being trained on unlabelled data alone, performs lower than our method which additionally leverages the potential of labelled sketch-photo pairs.
}

\vspace{-0.1cm} 
\subsection{Ablation Study}
\vspace{-0.2cm}

\keypoint{Importance of loss objectives:} {
To justify each loss in our network of a stronger baseline, we evaluate them in a strip-down fashion (\Cref{tab:abla}). Performance against cross-modal objective alone (Type-I), increase on aid from intra-modal objectives (Type-II), as the latter encourages distancing of multiple photos visually close to the sketch, based on finer intra-modal relative discrimination. 
Adding EMA (Ours) optimisation \cite{yaz2018unusual} ensures smoother convergence of above objectives with a slight increase (by $2.96\%$) in accuracy.
In similar spirit we perform experiments on losses of distillation paradigm (Eqn. \ref{equ:Distillation_student}). Performance using unlabelled data only (Type-III) rises with added knowledge from labelled photos (Type-IV). However, without aligning sketches using $\mathcal{L}_\text{KL}^{s_L}$ it lags behind \textit{Ours-Full} by 1.14\%.
}

\begin{table}[H]
\footnotesize
\setlength{\tabcolsep}{2.5pt}
\renewcommand{\arraystretch}{0.8}
\centering
\vspace{-0.4cm}
    \centering
    \caption{\small{Ablative study on QMUL-ShoeV2}}
    \vspace{-0.3cm}
    \begin{tabular}{cccccccc}
        \toprule
        \\[-0.25cm]
        Type & $\mathcal{L}^\text{CM}_\text{Tri}$ & $\mathcal{L}^\text{IM}_\text{Tri}$ & EMA 
        & $\mathcal{L}_\text{KL}^{p_U}$ & $\mathcal{L}_\text{KL}^{p_L}$ & $\mathcal{L}_\text{KL}^{s_L}$
        & Top-1 (\%)  \\[+0.05cm]                        
        \midrule\\[-0.25cm]
        I    & \checkmark &      -     &      -     & \checkmark & \checkmark & \checkmark &  43.28 \\ 
        II   & \checkmark & \checkmark &      -     & \checkmark & \checkmark & \checkmark &  45.39 \\ 
        III  & \checkmark & \checkmark & \checkmark & \checkmark &      -     &      -     &  46.50 \\ 
        IV   & \checkmark & \checkmark & \checkmark & \checkmark & \checkmark &      -     &  47.21 \\ 
   Ours-Full & \checkmark & \checkmark & \checkmark & \checkmark & \checkmark & \checkmark &  \bf48.35 \\ 
        \bottomrule
    \end{tabular}
    \label{tab:abla}
\vspace{-0.35cm}
\end{table}

\keypoint{Augmentation Strategy:} {We design a few experiments to explore other photo augmentation techniques~\cite{abayomi2021cassava} for $\mathcal{L}^{\text{IM}_p}_\text{Tri}$ (Eqn. \ref{equ:intramodal}) like, colour distortion, partial blurring and random shift in sharpness. Outperforming their respective scores of 46.14\%, 47.32\% and 47.54\% (Acc@1-ShoeV2) our methods confirms our intuition that morphological augmentations would better direct the loss objective, as it instills discrimination based on information exclusive to a sketch. }

\vspace{-0.0cm}
\keypoint{Influence of Distillation Token ($\Delta$):}
To justify its importance we explore two more designs. (A) Without any explicit distillation token we perform knowledge distillation using the same feature used for discriminative learning. (B) Following \cite{touvron2021training} we append a learnable token to the input of the last stage, which after processing acts as the distillation feature.
Although case-B ($45.31\%$) surpasses case-A ($44.71\%$) by $1.6\%$ in Acc@1 on QMUL-ShoeV2, confirming the need of a dedicated distillation token, it lags behind by $3.04\%$ to ours. We argue that our design of engaging the distillation token at every level via residual connection instills the inductive bias modelled by the pyramidal down-scaling, into the distillation token via transformer layers, thus creating a better representation.

\keypoint{Further Analysis:} { %
{(i) Although ours is a larger model than the earlier SOTA of Triplet-SN \cite{yu2016sketch}, it offers $\approx$20-25\% gain while taking similar inference time (0.37ms/ShoeV2). 
(ii) Our method takes 44.1 ms per training step compared 56.83 ms of SOTA StyleMeUP\cite{sain2021stylemeup} , proving itself as a strong and efficient method. 
(iii) Although our method takes more hyperparameters than the simpler Triplet-SN\cite{yu2016sketch} SOTA, they are quick to tune, and justifies itself with a boost of $\approx$25\% Acc@1 on Chair-V2 \cite{yu2016sketch}. 
}
}

\vspace{-0.15cm}
\subsection{Multi-category FG-SBIR via Unlabelled Photos}
\vspace{-0.1cm}


\noindent 
{Our pilot study (\cref{sec:pilot}) reveals FG-SBIR models to perform poorly on classes lacking sketch-photo pairs. This has been explored in a few recent FG-SBIR works. Apart from \textit{Jigsaw-SN} \cite{pang2020solving} (\cref{sec:competitors}), (\emph{CC-Gen})  \cite{pang2019generalising} takes a cross-category (CC) domain-generalisation approach, modelling a universal manifold of prototypical visual sketch traits that dynamically embeds sketch and photo, to generalise on unseen categories.
Recently, \cite{bhunia2022adaptive} uses a dedicated meta-learning framework that adapts a trained model to new classes using a few corresponding sketch-photo pairs as support. However, both methods have access to sufficient \cite{pang2020solving} or few \cite{bhunia2022adaptive} sketch-photo pairs of novel classes, unlike our setup of \textit{no paired sketches}. This goes beyond the dependency on paired sketches offering a realistic mid-ground between \textit{standard} \cite{yu2016sketch} and zero-shot \cite{dey2019doodle} inference setup, where \textit{not all} classes have paired sketches, but their photos can be collected \textit{easily}. Consequently,  following \cite{yelamarthi2018zero} we split Sketchy \cite{sangkloy2016sketchy} as 21 unseen-test classes, and 104 training classes with a 70:30 training:validation split in the latter. {Unlike existing methods, our student additionally trains via distillation from photos of 21 unseen classes for better retrieval accuracy.} For retrieval a separate gallery is maintained per category, and average top-q accuracy across 21 classes is reported (\Cref{tab:cross-category}). We compare with a few methods that leverage data of novel classes, besides labelled data for training: \textit{Jigsaw-SN} (extending \cite{pang2020solving}) trains via an auxiliary jigsaw-solving task on unlabelled photos without any fine-tuning (our setup). \textit{Adaptive-SN}~\cite{bhunia2022adaptive} trains on a few sketch-photo pairs of new classes. \textit{B-EdgePretrain} and \textit{B-Edge2Sketch} trains on edge-maps and synthetic sketches via \cite{riaz2018learning} from edge-maps of unlabelled photos respectively. \textit{CC-Gen}\cite{pang2019generalising} evaluates in zero-shot \cite{dey2019doodle} setup with \textit{no data} from novel classes. Consequently \textit{CC-gen} scores lower (\Cref{tab:cross-category}), compared to \textit{Jigsaw-SN} with its auxiliary jigsaw-solving task, whereas baselines using edge-maps of unlabelled images as \textit{pseudo}-sketches score relatively better. {Even without using real sketch-photo pairs of test-set for quick-adaptation like \textit{Adaptive-SN}~\cite{bhunia2022adaptive}, we achieve a competitive retrieval accuracy (2.47\% Acc@1).}
}

\vspace{-0.30cm}
{\setlength{\tabcolsep}{2.5pt}
\renewcommand{\arraystretch}{0.8}
\begin{table}[!hbt]
    \centering
    \footnotesize 
    \caption{\footnotesize{Cross-category FG-SBIR on Sketchy \cite{sangkloy2016sketchy}.}}
    \vspace{-0.20cm}
    \label{tab:cross_cat}
    \begin{tabular}{lcc|lcc} 
    \toprule
    \multirow{2}{*}{Methods} & \multicolumn{2}{c|}{Sketchy (\%)} &
    \multirow{2}{*}{Methods} & \multicolumn{2}{c}{Sketchy (\%)} 
    \\ \cmidrule(lr){2-3}  \cmidrule(lr){5-6}
    & Top-1 & Top-5 & & Top-1 & Top-5 \\
    \cmidrule(lr){1-3}  \cmidrule(lr){4-6}
    Jigsaw-SN \cite{pang2020solving}      & 23.16  & 44.63  & B-Edge-Pretrain & 24.81  & 46.24 \\
    Adaptive-SN \cite{bhunia2022adaptive} & 32.71  & 53.42  & B-Edge2Sketch   & 25.74  & 48.36 \\
    CC-Gen~\cite{pang2019generalising}    & 22.73  & 42.32  & \bf Ours-Full       & \bf 30.24  & \bf 51.65 \\
    \bottomrule
    \end{tabular}
    \label{tab:cross-category}
    \vspace{-0.30cm}
\end{table}
}

\vspace{-0.3cm}  
\section{Conclusion}
\vspace{-0.2cm}  
\noindent 
In this paper we put forth a strong baseline for FG-SBIR with PVT\cite{wang2021pyramid}-backbone, and offer a novel paradigm that at its core aims at learning from unlabelled data in FG-SBIR by distilling knowledge from unlabelled photos. 
While our proposed intra-modal triplet loss increases feature separation in model's latent space, an EMA paradigm stabilises its training. Importantly, we for the first time introduce a distillation token in PVT architecture that explicitly caters to knowledge distillation. Extensive experiments against existing frameworks and various baselines show our method to outperform them, thus proving its significance.

{\small
\bibliographystyle{ieee_fullname}
\bibliography{main}
}

\cleardoublepage

\renewcommand{\thefootnote}{\fnsymbol{footnote}}
\makeatletter
\renewcommand\@makefnmark{\hbox{\@textsuperscript{\normalfont\color{black}\@thefnmark}}}
\renewcommand\@makefntext[1]{%
  \parindent 1em\noindent
            \hb@xt@1.8em{%
                \hss\@textsuperscript{\normalfont\@thefnmark}}#1}
\makeatother
\onecolumn{
\begin{center}
\title{\vspace{+0.5cm}\Large{\textbf{Supplementary material for \\Exploiting Unlabelled Photos for Stronger Fine-Grained SBIR \\}}}
\vspace{+0.6cm}
\author{
    Aneeshan Sain\textsuperscript{1,2}  \hspace{.2cm}
    Ayan Kumar Bhunia\textsuperscript{1} \hspace{.3cm}
    Subhadeep Koley\textsuperscript{1,2}  \hspace{.2cm}
    Pinaki Nath Chowdhury\textsuperscript{1,2}  \hspace{.2cm}\\
    Soumitri Chattopadhyay\footnote{Interned with SketchX}  \hspace{.2cm}
    Tao Xiang\textsuperscript{1,2}\hspace{.2cm}  
    Yi-Zhe Song\textsuperscript{1,2} \\
    \textsuperscript{1}SketchX, CVSSP, University of Surrey, United Kingdom.  \\
    \textsuperscript{2}iFlyTek-Surrey Joint Research Centre on Artificial Intelligence.\\
    {\tt\small \{a.sain, a.bhunia, p.chowdhury,  t.xiang, y.song\}@surrey.ac.uk} 
}
\end{center}
\vspace{+0.2cm}
}

\renewcommand\thesection{\Alph{section}}
\setcounter{section}{0}

\section{Alternative to Triplet Loss} 
Factually, for photos, intra-modal triplet loss is a self-supervised objective. For a cross-modal problem such as ours however, triplet loss can offer some leeway in better conditioning the joint photo-sketch embedding \cite{yu2016sketch,song2017deep,bhunia2020sketch}. Empirically, when compared with contrastive loss \cite{chaudhari2019attentive} as a self-supervised objective (while keeping everything else the same), triplet loss performs better (45.68\% on ShoeV2) further justifying our case.

\section{On the need for Knowledge Distillation}
During student-training, three objectives are learnt -- cross-modal separation, intra-modal separation between photos and that between sketches. As learning everything together is difficult, we decouple the process by first training a teacher completely with intra-modal triplet loss on photos, and then use the trained teacher's photo discrimination knowledge to better guide the student (FG-SBIR) during its training. 

\section{On using additional datasets} 
TU-Berlin and QuickDraw are sketch-only datasets designed towards sketch classification. Some \cite{collomosse2019livesketch,dey2019doodle} did augment them for \textit{category-level }SBIR \cite{bui2018deep,dutta2019semantically}, by sourcing unpaired photos. These however do not work for our instance-level setting - we need instance-level sketch-photo correspondences. The idea of abstraction-influence is very interesting, which shall be considered as a future work. 

\section{Dealing with scarcity of sketch-data} 
Distilling from unlabelled photos is beneficial as they are abundantly available, unlike sketches that require time and human effort to collect \cite{bhunia2021more}. On distilling from only sketches there is minimal increment from teacher supervision (44.51\% vs. 44.18 \% on ShoeV2) as compared to that from photos (48.35\% vs. 44.18\% on ShoeV2). Faithful sketch-generation in photo-to-sketch generation tasks is challenging; it's difficult to quantify its generation-quality, and they hardly generalise to human-sketches \cite{bhunia2021more}. Using CLIPasso \cite{vinker2022clipasso} for sketch-generation instead of teacher-supervision, hence delivers a poor result of 38.57\% on ShoeV2. Although works have explored augmenting sketches via stroke-dropping/deformations \cite{yu2016sketch}, or as line-drawings \cite{chan2022learning}, resulting sketches mostly follow edge-maps thus being less reliable. On using \cite{chan2022learning} instead of teacher-supervision, we obtained a poorer result of 39.23\% compared to our 48.35\% on ShoeV2, thus proving our method to be simpler and more efficient.

\section{Clarity on training teacher} The teacher comprises an ImageNet pre-trained PVT backbone trained on 60,502 additional photos from Sketchy (ext) \cite{liu2017deep} for Sketchy; 50,025 photos of UT-Zap50k \cite{yu2014fine} for ShoeV2 \cite{yu2016sketch}; and 7,800 photos from websites like IKEA, etc \cite{pang2020solving} for ChairV2 \cite{sain2020cross}.

\section{Optimisation for multi-task objectives} 
We used the available toolbox of \textit{WandB Sweeps} for quick tuning of hyper-parameters, which provided 48.35\% Acc@1 on ShoeV2. Even on using complex loss balancing approach of \cite{groenendijk2021multi}, we obtain a close 47.94\%. Furthermore, changing the hyper-parameter values by $\pm10\%$, causes a mere $\pm0.5\%$ change in Acc@1 on ShoeV2 \cite{yu2016sketch}. This proves that our method despite needing five loss objectives, is quick to tune, thus being easily reproducible.

\section{Clarity on training stability} 
\noindent $\bullet$ Learning rate decay: We used exponential rate decay with initial learning rate of 0.001 and decay factor of 0.2.\\
$\bullet$ Large batch-size: We used 256 batch-size via gradient accumulation \cite{patel2022recall} on 4 V100-GPU machines.\\
$\bullet$ Reducing augmentations: We used augmentation (random horizontal flipping only) on just 30\% of training data.\\
Plots below show the above methods' implementation on top of CNN-Baseline. Least gittering in \textit{Ours} shows our EMA approach to be superior.

\begin{figure}[!hbt]
    \includegraphics[width=\linewidth]{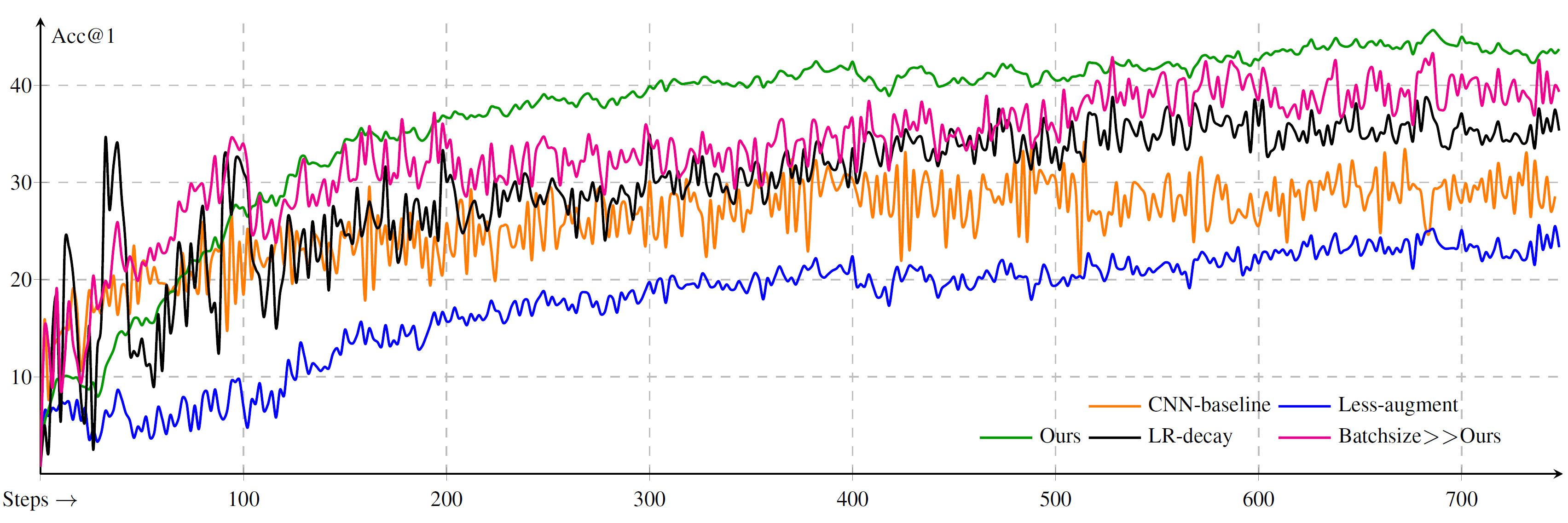}
    \vspace{-0.6cm}
    \caption*{Evaluation accuracy at every $100^\text{th}$ training-step [Best if zoomed]. 
   }
   \label{fig:plot_instability}
   \vspace{-0.4cm}
\end{figure}

\section{Further clarity on experimental results}
As we intended to show PVT \cite{wang2021pyramid} is a better backbone than the earlier CNN-based ones, we compared prior state-of-arts to our method using different backbones. Furthermore, for the methods having code available, we replaced their backbones with PVT, only to obtain inferior results (32.68\% for \cite{yu2016sketch} and 34.12\% for \cite{song2017deep}), thus proving ours as better. Furthermore,  our method surpasses by 8.33\% on ShoeV2, against a contemporary method of TC-Net \cite{lin2019tc}, despite having a lesser complexity of training and simpler loss objectives than the latter.




\end{document}